\documentclass[10pt,twocolumn,letterpaper]{article}
\usepackage[pagenumbers]{cvpr}

\DeclareRobustCommand{\todo}[1]{
  \ifmmode\text{\color{red}[\ifblank{#1}{TODO}{#1}]}
  \else{\color{red}[\ifblank{#1}{TODO}{#1}]}\fi
  \PackageWarning{gjbench}{Unresolved \string\todo}}

\newcommand{\bench}{GJBench\xspace}

\usepackage{multirow}

\usepackage{algorithm}
\usepackage[noend]{algpseudocode}

\definecolor{cvprblue}{rgb}{0.21,0.49,0.74}
\usepackage[pagebackref,breaklinks,colorlinks,allcolors=cvprblue]{hyperref}
\hypersetup{pdfauthor={Jhen-Ke Lin, Chung-Chun Wang}, pdftitle={Can Image Generators Hold a Scene Together When Space Is Broken Apart?}}
\def\confName{CVPR}
\def\confYear{2027}
\title{Can Image Generators Hold a Scene Together When Space Is Broken Apart?}
\author{Jhen-Ke Lin \qquad Chung-Chun Wang\\
National Yang Ming Chiao Tung University\\
{\tt\small \{jacob.cs14, takalawang.cs14\}@nycu.edu.tw}}
\begin{document}
\maketitle
\begin{abstract}
Image generators are good at drawing scenes in which things that belong together appear next to each other.
In almost every image they generate, and in almost every benchmark used to evaluate these images, ``belonging to the same whole'' and ``adjacent to each other'' go together.
Therefore, we cannot know whether the model truly grasps the scene as a whole, or simply makes each part connect with its adjacent parts.
We propose \bench to disentangle the two.
It requires the model to draw a scene as a scattered jigsaw puzzle: an image is cut into interlocking pieces, then moved apart and rotated.
Thus, the spatial plausibility of the scene exists only in how these pieces collectively point to a whole.
The model may have a reference image, a target layout, both, or neither.
We separately evaluate three requirements: whether the scene is broken down into a specified number of separate pieces, whether the pieces interlock, and whether the pieces can reconstruct the reference scene.
While every open and closed model we evaluated could draw these scenes well as a whole, none could keep them together as scattered puzzle pieces, and the three requirements categorized failures into different patterns.
The models draw separate pieces but miscount them in different ways, and unless their outlines are provided, the pieces rarely interlock, and their content does not reconstruct the reference image.
We also found that the target layout helps the puzzle pieces fit together, while the reference image improves neither their count nor their fit.
These findings show that keeping a scene continuous across space that the canvas no longer shows remains an open problem for image generators.
\end{abstract}

\newcommand{\gjbteaser}[2]{\includegraphics[height=2.02cm]{generated/teaser/#1--geometry-04--#2.jpg}}
\begin{figure*}[t]
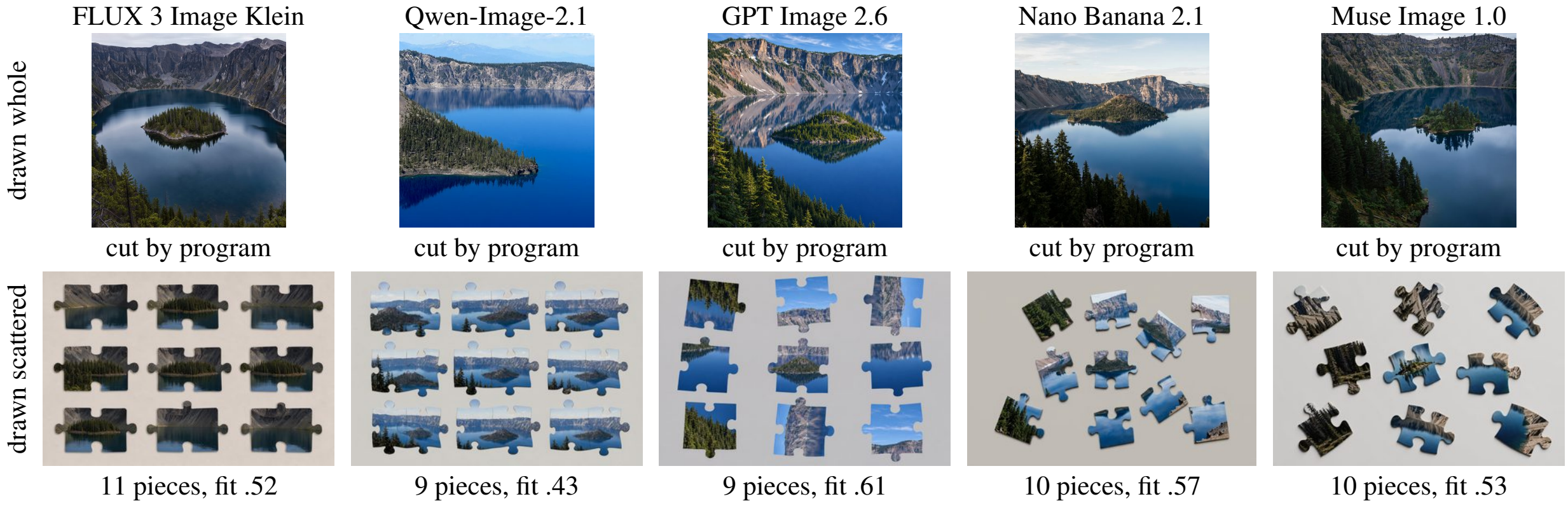

  \centering
  \footnotesize
  \setlength{\tabcolsep}{2.5pt}
  \renewcommand{\arraystretch}{0.9}
  \begin{tabular}{@{}r ccccc@{}}
    & FLUX 3 Image Klein & Qwen-Image-2.1 & GPT Image 2.6 & Nano Banana 2.1 & Muse Image 1.0 \\
    \rotatebox{90}{\parbox{2.02cm}{\centering drawn whole}} &
    \gjbteaser{flux3-klein}{whole--whole} & \gjbteaser{qwen-image-2.1}{whole--whole} &
    \gjbteaser{gpt-image}{whole--whole} & \gjbteaser{gemini-image}{whole--whole} &
    \gjbteaser{muse-image}{whole--whole} \\
    & cut by program & cut by program & cut by program & cut by program & cut by program \\[2pt]
    \rotatebox{90}{\parbox{2.02cm}{\centering drawn scattered}} &
    \gjbteaser{flux3-klein}{9--no_ref_no_layout} & \gjbteaser{qwen-image-2.1}{9--no_ref_no_layout} &
    \gjbteaser{gpt-image}{9--no_ref_no_layout} & \gjbteaser{gemini-image}{9--no_ref_no_layout} &
    \gjbteaser{muse-image}{9--no_ref_no_layout} \\
    & 11 pieces, fit .52 & 9 pieces, fit .43 & 9 pieces, fit .61 & 10 pieces, fit .57 & 10 pieces, fit .53 \\
  \end{tabular}
  \caption{\textbf{Every system draws the picture, but none scatters it as a jigsaw.}
    Five generators receive the same brief (a forested island in a crater lake).
    Top: each system's picture drawn whole, which a program then cuts into nine interlocking pieces.
    Bottom: the same system asked for the nine pieces scattered on a plain background, from text alone.
    Labels give the pieces found and the interlock score (\cref{sec:evaluation}). Each exact cut above has 9 pieces and scores .94. The scene is chosen by a fixed rule.}
  \label{fig:teaser}
\end{figure*}

\section{Introduction}
\label{sec:intro}

Image generators now render complete scenes in convincing detail and follow long compositional prompts~\cite{rombach2022high,esser2024scaling,wu2025qwenimage}.
A coherent scene is more than plausible local texture: surfaces and contours continue, objects rest on their supports, and objects keep sensible relations to one another.
In almost every image a generator produces, however, the parts of one scene are also neighbors on the canvas, and the same holds in almost every benchmark that scores such images.
It therefore remains unclear whether a generator holds a scene together as a whole or only keeps each neighborhood consistent with the next.

Putting a picture together from its parts has long served as a probe of visual understanding. Children's puzzle play predicts their later spatial transformation skill~\cite{levine2012early}. Networks that learn to solve jigsaws learn object parts and how image regions relate in space~\cite{noroozi2016unsupervised,carlucci2019domain}, and the same training improves the spatial understanding of multimodal models~\cite{wang2025jigsawr1,wu2026visual}.
A solved jigsaw, however, ends as one picture whose parts are neighbors on the canvas, so matching each edge with the next can suffice~\cite{doersch2015unsupervised,noroozi2016unsupervised}.
Drawing a scattered jigsaw reverses the task. The generator must decide what each piece shows and where it continues, with no neighbor on the canvas to match.

Scattered pieces thus pull apart the two properties that the canvas usually joins. They lie apart, yet they belong to one picture, which continues only across seams that the canvas no longer shows.
Perception draws the same distinction: regions divided by a gap count as separate objects~\cite{spelke1990principles}, yet they are seen as one object when their edges and surfaces continue across the gap~\cite{kellman1991theory,yin1997surface}.
We therefore ask a generator for one image of pieces, scattered and rotated on a plain background, that fit back together into a picture whose content continues across every seam (\cref{fig:teaser}). The two halves of a bridge may land at opposite ends of the canvas, yet put back together they must form one bridge.
Without a reference, this picture appears in neither the input nor the output. Editing, multi-panel generation and panoramas always show it in one of the two (\cref{sec:related}).

Such an image must meet three requirements, and \bench scores each as a separate question, as intuitive-physics benchmarks test each principle on its own~\cite{piloto2022intuitive}.
\emph{Scatter} asks whether the image shows the requested number of pieces separated by background, a counting check like those of compositional benchmarks~\cite{ghosh2023geneval}.
\emph{Interlock} asks whether the shapes of the pieces fit together without overlap or gap in one arrangement, the property that solvers working from shape alone rely on~\cite{freeman1964apictorial,goldberg2004global}.
\emph{Reconstruction} asks whether the pieces, put back in place, restore the reference when one is given. Jigsaw solvers are scored by the same standard~\cite{cho2010probabilistic,gallagher2012jigsaw}, and we grade it with structural similarity~\cite{wang2003multiscale} and with precision and recall in the form of panoptic quality~\cite{kirillov2019panoptic}.
Continuity across a seam is a low-level cue, even when the generator must produce it. We therefore treat it as a necessary condition for a coherent output, not as evidence of understanding.
The evaluation is deterministic, reads only pixels and fits none of its constants to generated outputs. It uses no learned judge, because multimodal models overlook fine visual detail~\cite{tong2024eyes,rahmanzadehgervi2024vision}, as a small pilot confirms (appendix).

We apply \bench to open-weight and proprietary generators, with and without a reference picture and a target layout. Each generator also draws every scene whole, and a program cuts this picture exactly into pieces.
These cuts score near the top of every question, so every system draws the scenes well. Yet no system keeps a scene together once it must spread the scene across separate pieces (\cref{sec:results}).
The three questions show that each system fails in its own way. All of them draw separate pieces, although they miscount in different ways. Their pieces seldom interlock unless a layout shows the outlines, and their content does not restore the reference piece by piece.
A layout helps the pieces fit together, but a reference helps neither their count nor their fit.

\noindent In summary, we contribute:
\begin{itemize}[leftmargin=*,itemsep=1pt,topsep=2pt]
  \item \textbf{A task that tests spatial coherence when a scene is broken apart}: scattered jigsaw generation, which separates where content sits on the canvas from where it belongs in the scene and, without a reference, asks for the pieces of a scene given only as text.
  \item \textbf{A deterministic evaluation with three separately scored questions}, scatter, interlock and reconstruction, each built from established measures of counting, puzzle assembly and image similarity and checked on constructed outputs with known answers.
  \item \textbf{A controlled study across generator families} that crosses a reference with a layout, varies the piece count and adds a control that cuts each model's own whole picture. Prompts, layouts, evaluation code and per-output records will be released.
\end{itemize}

\section{Related Work}
\label{sec:related}

\paragraph{Evaluating generated images.}
Compositional benchmarks test whether an image contains the objects, attributes, counts and relations a prompt names, scored with detectors~\cite{ghosh2023geneval}, task-specific metrics~\cite{huang2023t2icompbench} or visual question answering~\cite{hu2023tifa,cho2024davidsonian,lin2024evaluating}.
Editing benchmarks also require unedited content to survive~\cite{zhang2023magicbrush,ku2024imagenhub}, and ABACUS verifies this preservation instance by instance against panoptic ground truth~\cite{lassoued2026abacus}.
Learned evaluators built on embeddings~\cite{hessel2021clipscore}, promptable segmentation~\cite{carion2025sam3} or vision--language judges~\cite{ku2024viescore} scale such checks, but they share the blind spots of the models they build on. VLM-based faithfulness metrics fail to significantly outperform CLIPScore at ordering images by error count~\cite{saxon2024who}, multimodal LLMs overlook fine visual detail~\cite{tong2024eyes,rahmanzadehgervi2024vision}, and multimodal judges show biases and uneven adherence to stated criteria~\cite{chen2024mllm,xiong2026multicrit}.
Each of these properties can be checked within one region of the image.
Whether pieces fit back into one picture, by contrast, depends on regions far apart on the canvas, and \bench measures it from pixels alone.

\paragraph{Jigsaws in generation.}
Reasoning benchmarks use jigsaws in the solving direction: a model restores shuffled tiles or fills a missing tile of a given image and is scored against the original~\cite{li2026girbench,han2026vigor,zou2026unimmmu}.
Dedicated methods design puzzles whose cut the method itself fixes, by drawing the pieces by hand and applying them as a pixel permutation while denoising one assembled picture~\cite{geng2024visual} or by optimizing the outlines of the pieces~\cite{chew2026whimsy}.
Chew \etal~\cite{chew2026whimsy} also prompt general-purpose models directly for jigsaws and observe, qualitatively, limited geometric control and pieces that lack consistency.
We turn this direct setting into a measured benchmark. The model draws all pieces in a single image, places them itself when no layout is given, and invents the picture when no reference is given.

\paragraph{Coherence across views and panels.}
Joint diffusion fuses overlapping windows into a panorama~\cite{bartal2023multidiffusion,lee2023syncdiffusion}, MVDiffusion generates overlapping views of one panorama through correspondence-aware attention~\cite{tang2023mvdiffusion}, and multi-view diffusion carries this coupling to 3D objects~\cite{shi2024mvdream}.
Story and multi-panel methods keep one subject consistent across panels~\cite{zhou2024storydiffusion,tewel2024training}, and ImageTime asks for the ordered states of one process within one image~\cite{wu2026imagetime}.
In these settings the parts overlap, sit at known positions, or each form a complete picture, so their agreement can be checked on shared pixels or within each part. Separate jigsaw pieces share only their seams (\cref{sec:task}).

\paragraph{Jigsaw puzzle solving.}
The first solvers assembled puzzles from the shape of the pieces alone~\cite{freeman1964apictorial,goldberg2004global}.
Solvers for square pieces with content score how well the content continues across a candidate seam and assemble the pieces greedily, probabilistically, with genetic algorithms or through loop consistency, also when orientation is unknown~\cite{cho2010probabilistic,pomeranz2011fully,gallagher2012jigsaw,sholomon2013genetic,son2014solving,paikin2015solving}. Later work handles polygonal cuts, eroded fragments, learned compatibility, diffusion-based reassembly~\cite{harel2021crossing,derech2021solving,shahar2026missing,alali2026fix,scarpellini2024diffassemble,liu2024solving} and interlocking tabs in VLM puzzle benchmarks~\cite{li2026jigshape}.
\emph{Best buddies}, two sides that each rank the other as their most compatible partner~\cite{pomeranz2011fully}, are rarely wrong, while smooth regions make every candidate look alike~\cite{paikin2015solving}.
Solvers assume that the pieces come from a real picture and search for their arrangement. \bench asks whether generated pieces form a puzzle and come from one picture at all.

\paragraph{Scoring reassembly.}
Puzzle solvers are scored against the true arrangement, by the share of pieces in their correct cell or of correctly joined neighbors~\cite{cho2010probabilistic,gallagher2012jigsaw,yu2016solving}. Fracture reassembly reports the share of correctly reassembled parts beside rotation and translation errors~\cite{sellan2022breaking}.
Panoptic segmentation matches predicted and true segments one to one and, like 3D reconstruction, combines precision and recall into one score~\cite{kirillov2019panoptic,tatarchenko2019single}. Editing benchmarks measure what a mask leaves unchanged with pixel and structural similarity~\cite{ju2024pnp,wang2004image}.
\bench combines these practices. A generator states neither where its pieces go nor which part of the picture each one shows, so the evaluation recovers both before scoring (\cref{sec:evaluation}).

\paragraph{Generators.}
Current open-weight and proprietary generators take text and image inputs and both generate and edit images~\cite{bfl2025flux2,bfl2026flux2klein,bfl2026flux3image,wu2025qwenimage,qwen2026qwenimage21,openai2026gptimage26,google2026nanobanana21,meta2026museimage}. We test five of them in all four conditions through their own interfaces (\cref{sec:experiments}).

\section{Scattered Interlocking Jigsaw Generation}
\label{sec:task}

Each request gives a text brief and a piece count $N=n^2$ with $n\in\{2,3,4\}$.
The generator returns one image that shows only a plain, uniform background and exactly $N$ pieces at the same scale, each fully surrounded by background and turned by a multiple of $90^\circ$.
The pieces come from an $n\times n$ jigsaw of square cells: outer sides are straight, and every shared side carries one tab and the matching socket.
To reassemble the puzzle, each piece may be moved and rotated but not mirrored, cropped, stretched or repainted. The content of a piece thus keeps the orientation and scale of the cell it was cut from.
An output succeeds when its pieces fit together into one picture whose content continues across every seam.
The generator decides which part of the picture each piece shows, where each piece lies and, without a reference, what the picture shows.

\begin{table}[t]
  \centering
  \footnotesize
  \setlength{\tabcolsep}{3pt}
  \begin{tabular}{@{}l l p{0.56\columnwidth}@{}}
    \toprule
    Condition & Images & Text appended to the shared prompt \\
    \midrule
    \textsc{Text} & none & \texttt{Place and rotate the pieces at random across the gray background.} \\
    \textsc{Ref} & reference & \texttt{Image 1 shows the scene exactly.}\ + the \textsc{Text} clause \\
    \textsc{Layout} & layout & \texttt{Each piece takes the exact place of a different piece shape in Image 1, on Image 1's background.} \\
    \textsc{Ref+Layout} & both & the \textsc{Ref} clause + the \textsc{Layout} clause, naming Image~2 \\
    \bottomrule
  \end{tabular}
  \caption{\textbf{The four conditions differ only in what is appended} to one shared prompt, brief included (full prompt in the appendix).}
  \label{tab:cells}
\end{table}

We cross two optional inputs (\cref{tab:cells}).
A \emph{reference} is an image of the whole scene, and a \emph{layout} shows unlabeled piece outlines at the target positions and rotations.
\textsc{Text} gives neither and asks the most of a generator, which must invent the picture and scatter it on its own.
\textsc{Ref} asks whether a given picture survives being scattered, and \textsc{Layout} asks whether a given arrangement helps keep an invented picture coherent.
Every prompt starts with the same text, so the difference between two conditions measures the full effect of adding an input, including any change in how a system handles image inputs.
The shared text ties the brief to the reassembled picture (\texttt{When fitted together, they form one continuous scene:~\{brief\}}).
We hold every brief to fixed rules: every countable object carries a number word, relations refer to the upright reassembled picture, no word refers to the puzzle itself, such as \emph{piece}, and no text appears in the scene.
Each brief is reused word for word across conditions, piece counts and the whole-scene control below.
Scenes fall into four families, each testing a different way in which content crosses a seam: \emph{lines and topology} (roads, rivers and bridges that run through many seams), \emph{object parts} (one object divided among pieces), \emph{multi-object relations} (counts and relative positions spread across pieces), and \emph{scene geometry and lighting} (horizon, perspective, material and shadow).

Success thus rests on three requirements that can fail independently: the image must show $N$ separate pieces, the pieces must interlock as one jigsaw, and their content must form one picture.
The evaluation scores each one separately (\cref{sec:evaluation}).
The optional inputs change what the generator sees but never what the evaluator sees. The evaluator reads the output image and $N$. It reads the reference only to score the third requirement in the conditions that provide one. It never reads the prompt, the layout, the condition or the identity of the model.

Every generator also draws each brief as one whole picture, from a prompt that never mentions pieces. A program then cuts this picture exactly with the benchmark's piece shapes and lays the pieces out on the required background.
This whole-scene control gives each scene and generator an output that meets every requirement by construction and carries the generator's own content.
Its scores mark the ceiling of every measure on that content. Comparing a generated output with its control therefore separates a generator that cannot draw the scene from one that can draw it but cannot spread it across pieces.

\section{Evaluation}
\label{sec:evaluation}

\begin{figure*}[t]
  \centering
  \includegraphics[width=\textwidth]{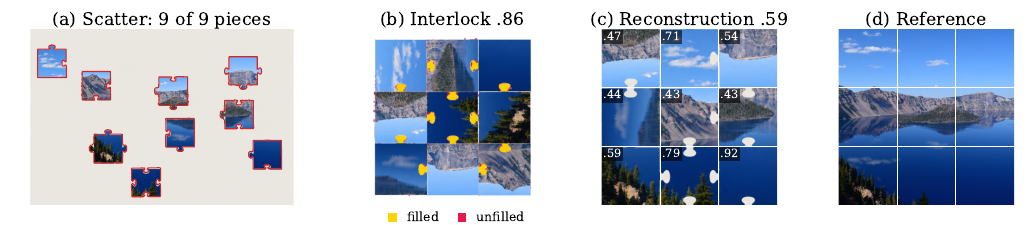}
  \caption{\textbf{One output read by the three questions.} (a) The pieces found on the background. (b) The best arrangement, with tab and socket pixels marked by whether the neighboring piece fills them. (c) Each piece on the reference cell it is paired with, labeled by its similarity to that cell. The output comes from GPT Image 2.6 with a reference and a layout, chosen by a fixed rule.}
  \label{fig:eval}
\end{figure*}

The three questions follow the three requirements of the task (\cref{sec:task}), and \cref{fig:eval} applies them to one output.
\emph{Scatter} (RQ1) asks whether the image shows the requested number of separate pieces on a plain background, each an object set apart from the others by a gap~\cite{spelke1990principles}.
\emph{Interlock} (RQ2) asks whether those pieces form a jigsaw: their shapes meet without overlap or gap, and every tab finds a matching socket in one arrangement.
\emph{Reconstruction} (RQ3) asks whether the pieces, moved back and turned upright, restore the picture they were cut from, so that the content of separate pieces joins into one whole~\cite{kellman1991theory}.
No score requires another to pass, so each can be read on its own. A generator that merges two pieces still receives an interlock score for the rest, and one whose pieces do not fit still receives a reconstruction score for their content.
Every step is deterministic and reads only the output image, $N$ and, for RQ3, the reference, which only the conditions with a reference provide.
Each definition either follows an established evaluation practice, cited where it is used, or is the plain reading of the task.

All three questions start by separating the pieces from the background.
The task asks for a plain background, so we take its color as the median of the outermost ring of pixels. Each pixel's distance from it is its largest difference over the color channels.
The histogram of these distances has one large background peak and a long tail of piece colors, the shape that the triangle threshold~\cite{zack1977automatic} was designed for.
Pixels at or below the threshold that connect to the image border form the background, and the pieces are the 8-connected components~\cite{rosenfeld1970connectivity} of the rest. A component counts as a piece only if some pixel lies more than two pixels inside its edge, which removes specks and thin lines.
Generated images contain compression noise and occasional soft shadows that this plain reading would mistake for pieces.
We therefore read each image in four fixed ways: as is, after a $3\times3$ median filter, with pixels that are darker versions of the background color counted as shadow, and with both.
Each question keeps its best score over the four readings. Trajectory benchmarks do the same when they align an estimate to the ground truth before measuring its error~\cite{sturm2012benchmark,horn1987closed}: noise that says nothing about the measured property is removed in the system's favor.
None of the readings splits, joins or reshapes a piece, so taking the best cannot hide the errors the questions measure. The same readings apply to every control and construction, whose scores they raise by about one hundredth, so the marks below barely move.

Scatter (RQ1) is answered by counting, as compositional benchmarks check that an image holds the number of objects a prompt names~\cite{ghosh2023geneval,huang2025t2icompbenchpp}.
An output has the exact count when it shows $N$ pieces, and its count error is the signed difference.
The count can hide two failures. Pieces drawn touching merge into one component, so a component about $m$ times the expected piece area, the total piece area divided by $N$, counts as $m-1$ merged pieces. Pieces that run off the canvas are cut off, which shows as a piece touching the image border.
RQ1 reports the share of outputs with the exact count, the merged pieces per output and the share of pieces cut by the border.

RQ2 and RQ3 both look at each piece on its own.
We crop each piece, rotate it upright with projection-profile skew estimation~\cite{bloomberg1995measuring}, and give it a \emph{core}, the square bounded by the four straight lines through its sides. Turning a piece upright by its shape leaves only its four quarter turns open.
On each side, the tab is the part of the piece outside the core line, and the socket is the background that cuts into the core from that line.
The core also sets the scale of the piece, because in an $n\times n$ cut it covers exactly one cell of the picture.

Interlock (RQ2) asks whether the pieces fit by shape alone, the problem studied by shape-only puzzle solvers~\cite{freeman1964apictorial,goldberg2004global}. Puzzle benchmarks for vision--language models score predicted neighbors by whether a tab meets a matching blank~\cite{li2026jigshape}.
\bench adds one requirement that makes the fit physical: all pieces must fit together in one $n\times n$ arrangement.
Two sides placed against each other meet at the midpoints of their core lines, as when two pieces are pushed together. Their \emph{fill} is the number of pixels where the tab of one lies inside the socket of the other, counted both ways.
An arrangement places the pieces in the grid cells in any order and any quarter turn. Its fill is the total over all inner seams, and the score is the Dice overlap~\cite{dice1945measures}
\begin{equation}
  \mathrm{RQ2}=\max_{\text{arrangements}}\frac{2\sum_{\text{seams}}\mathrm{fill}}{\sum_{\text{pieces, sides}}\bigl(|\text{tab}|+|\text{socket}|\bigr)}.
  \label{eq:rq2}
\end{equation}
A tab and socket that fill each other exactly add the same number of pixels to the numerator and the denominator. Tabs that face the border, tabs or sockets without a partner, and the sides of a piece left out of the grid add to the denominator alone.
Comparing every tab with every socket, however, would miss a common error. A tab shifted to one side would still find a socket shifted the same way on another piece, and in our tests such pieces scored like an exact cut.
In one arrangement, each piece can use only the neighbors that its cell gives it, so misplaced tabs lower the score.
We find the best arrangement exactly by solving a mixed-integer program with the HiGHS solver~\cite{huangfu2018parallelizing}, a formulation that jigsaw assembly has also been posed as~\cite{yu2016solving}. Most images take seconds, and a solve that reaches its time limit records its optimality gap.
RQ2 does not compare piece sizes, because real puzzle pieces differ in size and still fit, and RQ3 already ties each piece's content to its scale.

For reconstruction (RQ3), we start from how jigsaw solvers are scored: \emph{direct comparison} counts the share of pieces placed in their correct cell. When orientation is unknown, a piece counts only if it is also turned correctly~\cite{cho2010probabilistic,gallagher2012jigsaw}.
A generator, however, never states where its pieces belong. RQ3 therefore compares every piece with every cell of the reference and lets the best one-to-one pairing~\cite{kuhn1955hungarian} decide.
We lay each piece on a cell at the scale its core implies, with its tabs over the neighboring cells. We try it only in the four quarter turns that its shape leaves open, the same four orientations that self-supervised rotation prediction treats as recognizable from content~\cite{gidaris2018unsupervised}. Content turned or scaled inside the piece is therefore not corrected.
The similarity of piece and cell is MS-SSIM~\cite{wang2003multiscale} with the published scale weights and the Gaussian window of SSIM~\cite{wang2004image}, averaged over the color channels. It is computed over the piece's own pixels only, as editing benchmarks restrict similarity to the region an edit should leave unchanged~\cite{ju2024pnp}.
Two details keep the comparison fair to the piece. The reference is blurred and resampled to the piece's resolution, so no piece is asked for more detail than its pixels can hold. Outside the piece, the reference takes the piece's own pixels, so filter windows that cross the outline see the same surroundings in both images.
With $k$ the number of pieces found and $s_i$ the similarity of the $i$-th piece to its paired cell (zero for a piece left unpaired), the score takes the form of panoptic quality~\cite{kirillov2019panoptic},
\begin{equation}
  \mathrm{RQ3}=\frac{2\sum_i s_i}{N+k},\qquad
  \mathrm{P}=\frac{\sum_i s_i}{k},\qquad
  \mathrm{R}=\frac{\sum_i s_i}{N},
  \label{eq:rq3}
\end{equation}
the harmonic mean of a precision P and a recall R, which reconstruction benchmarks use to separate missing content from extra content~\cite{tatarchenko2019single}.
Missing pieces lower recall, a case that puzzle solvers also face~\cite{paikin2015solving}. Extra or duplicated pieces lower precision, since each can match only a cell that another piece already matches better.
Panoptic quality counts a pair only above an overlap of one half. RQ3 keeps every pair and weights it by its similarity, so the score falls gradually as an error grows.
One failure deserves its own name. Sometimes a piece shows the whole scene in miniature: it matches no cell but is clearly the scene, the difference between recognition and reconstruction pointed out by Tatarchenko \etal~\cite{tatarchenko2019single}. To detect it, we report beside RQ3 a diagnostic that compares each piece with the whole reference shrunk onto its core.

\begin{table}[t]
\centering
\footnotesize
\setlength{\tabcolsep}{3.5pt}
\begin{tabular}{@{}lcccccc@{}}
\toprule
 & \multicolumn{3}{c}{RQ2} & \multicolumn{3}{c}{RQ3} \\
\cmidrule(lr){2-4}\cmidrule(l){5-7}
Construction ($N$) & 4 & 9 & 16 & 4 & 9 & 16 \\
\midrule
Exact cut & .95 & .93 & .88 & .92 & .92 & .92 \\
One piece missing & .63 & .85 & .84 & .79 & .86 & .89 \\
Tabs shifted along their edge & .48 & .78 & .77 & \multicolumn{3}{c}{as exact} \\
Content turned $5^\circ$ in each piece & \multicolumn{3}{c}{as exact} & .67 & .75 & .80 \\
Content turned $20^\circ$ in each piece & \multicolumn{3}{c}{as exact} & .45 & .55 & .61 \\
Content at half scale in each piece & \multicolumn{3}{c}{as exact} & .36 & .46 & .52 \\
Every piece a thumbnail of the scene & \multicolumn{3}{c}{as exact} & .33 & .36 & .39 \\
Exact cut of another scene & \multicolumn{3}{c}{as exact} & .35 & .43 & .48 \\
\bottomrule
\end{tabular}
\caption{\textbf{Reference marks for reading RQ2 and RQ3.}}
\label{tab:anchors}
\end{table}

Neither similarity score runs from zero to one in practice, so we read both against constructed outputs whose answers are known (\cref{tab:anchors}). We build them from the benchmark's own scenes and layouts, and they mark the scale without setting any constant.
An exact cut marks a complete and correct reconstruction. It stays below one because tabs and sockets drawn on a pixel grid never fill each other perfectly and because the reference is resampled to each piece.
At the other end, an exact cut of another scene stands for pieces that contain nothing of the reference, since the structural similarity of unrelated images is not zero.
Between the two lie pieces damaged in known ways: a missing piece, and content turned or shrunk inside its outline.
Pieces that each show the whole scene in miniature score as low as another scene. The scene is recognizable, but none of the pieces contains a cell of the reference.
We read an RQ3 score by where it falls among these marks, which tells how much of the reference the pieces restore. The marks change with $N$, so we compare each result with the marks of its own $N$.
For RQ2, shifted tabs and a missing piece lower the score. Outlines swapped among the pieces of one cut still form a valid puzzle and keep the score of the exact cut.

\section{Experimental Setup}
\label{sec:experiments}

\begin{figure*}[t]
  \centering
  \includegraphics[width=\textwidth]{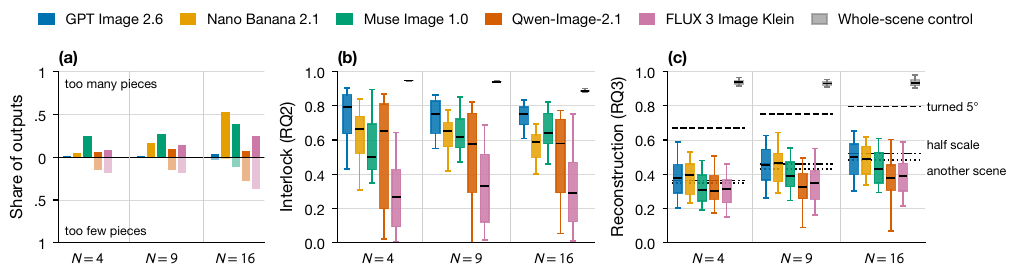}
  \caption{\textbf{Each system falls short in its own way.}
  (a) Share of outputs with too many pieces (up) or too few (down). The rest have exactly $N$.
  (b, c) Scores of single outputs. Boxes span the middle half of the outputs, whiskers the 5th to 95th percentile, and black bars mark the median.
  In (c), dashed and dotted segments are reference marks from \cref{tab:anchors}, drawn at each $N$. Gray boxes are the whole-scene controls.
  Each system has 128 outputs per $N$ in (a, b) and 64 in (c).}
  \label{fig:results}
\end{figure*}

\paragraph{Scenes.}
\bench uses 32 scenes, eight per content family (\cref{sec:task}).
Each scene is a Wikimedia Commons photograph admitted by these rules: a CC0 dedication with no further restriction; a square crop of at least $1024$ pixels, resized down to $1024\times1024$; no visible text, signs, logos or identifiable people; no near-duplicate among the scenes; and a crop border whose colors stay distinct from the required background.
LLM agents proposed candidates, crops and briefs under these rules. A separate agent cross-checked every scene against its crop, revising unsupported briefs and replacing any that broke a rule.

\paragraph{Systems.}
We test five systems, two open-weight and three proprietary.
FLUX 3 Image Klein~\cite{bfl2026flux3image} and Qwen-Image-2.1~\cite{wu2025qwenimage,qwen2026qwenimage21} run from fixed weight versions, seeded from their job identifiers.
As its model card requires, FLUX 3 Image Klein first passes each request through its prompt upsampler, a language model that reads the prompt and the condition images and rewrites them as one full scene caption. This caption is the only text its generator receives, and we keep it with every output.
The supplementary material adds FLUX.2 [klein] 4B~\cite{bfl2026flux2klein}, a smaller model from the same developer that runs without such a stage.
GPT Image 2.6~\cite{openai2026gptimage26}, Gemini Nano Banana 2.1~\cite{google2026nanobanana21} and Muse Image 1.0~\cite{meta2026museimage} are called through their providers' APIs, without a seed.
Every system accepts text alone, one image or two ordered images through its own interface, so every condition reaches every system without combining images into one. Each system runs at one fixed setting with no tools enabled (identifiers, settings, sizes and costs in the appendix).
Scattered jobs request a $1536\times1024$ canvas and whole-scene jobs a square one. Outputs are evaluated at the size returned.

\paragraph{Layouts.}
Each scene and piece count receives one layout: a $1536\times1024$ image of $N$ unlabeled piece outlines on the required background. The outlines are placed one at a time at random positions that keep a minimum gap from those already placed, with random quarter-turn rotations and the required piece scale.
The same layout serves all five systems, both layout conditions and the cut controls of that scene. Every layout comparison thus fixes the geometry.

\paragraph{Jobs.}
Crossing 32 scenes, four conditions, $N\in\{4,9,16\}$ and five systems gives 1{,}920 scattered jobs with one output each, plus 160 whole-scene pictures.
Each job keeps its first complete response: a refusal counts as the response, and only a network failure that returns nothing may be retried.

\paragraph{Whole-scene control.}
The layout program cuts each whole-scene picture exactly at every piece count and lays the pieces out in the scene's layout with anti-aliased edges on the required background. Exact cuts of the 32 reference crops give the same ceiling for the photographs themselves.

\paragraph{Measures.}
Every output receives the scatter and interlock scores, and every output in a condition with a reference receives the reconstruction score against the reference crop (\cref{sec:evaluation}). Jobs that return no image count as outputs with no pieces, so they have the wrong count and an interlock score of zero.
Each whole-scene control is scored the same way, with the generator's own picture as the reference, and so are exact cuts of the reference crops.
Each interlock solve has a budget of 60 seconds, extended to 600 seconds when the first budget runs out. Solves that still reach it report their best arrangement with the optimality gap, so their scores can only err low. Among the solves that set a score, this happens only at sixteen pieces (appendix).

\paragraph{Statistics.}
Every mean carries a 95\% interval from a bootstrap that resamples scenes, the unit shared by all outputs of a scene~\cite{field2007bootstrapping}, with 2{,}000 resamples.
We measure the effect of an input as the mean paired difference between outputs of the same system, scene and piece count that differ only in that input. It counts as a change when its interval excludes zero.
Each job yields one output, so the intervals describe variation over scenes, not over repeated samples.

\providecommand{\gjbbody}[1]{\csname @@input\endcsname generated/#1.tex }
\section{Results}
\label{sec:results}

\begin{table}[t]
\centering
\footnotesize
\setlength{\tabcolsep}{2.4pt}
\begin{tabular}{@{}lccccccccc@{}}
\toprule
 & \multicolumn{3}{c}{\shortstack{Exact count\\(RQ1)}} & \multicolumn{3}{c}{\shortstack{Interlock\\(RQ2)}} & \multicolumn{3}{c}{\shortstack{Reconstruction\\(RQ3)}} \\
\cmidrule(lr){2-4}\cmidrule(lr){5-7}\cmidrule(l){8-10}
$N$ & 4 & 9 & 16 & 4 & 9 & 16 & 4 & 9 & 16 \\
\midrule
\gjbbody{tab_v3_main_body}
\bottomrule
\end{tabular}
\caption{\textbf{No system meets all three requirements.} Means over all conditions, and for RQ3 over the two conditions with a reference. Control: whole-scene control. Read RQ2 and RQ3 against \cref{tab:anchors}. Intervals are in the appendix.}
\label{tab:main}
\end{table}

\begin{figure}[t]
  \centering
  \includegraphics[width=\linewidth]{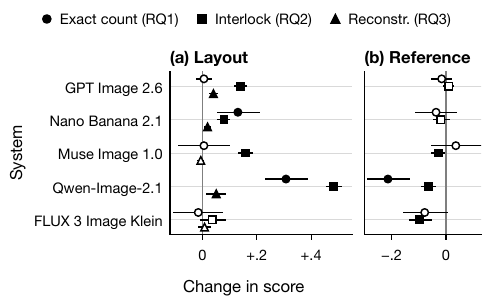}
  \caption{\textbf{A layout helps the pieces fit, and a reference does not.} Mean paired change in each score when an input is added, over scenes and piece counts, with 95\% scene-cluster bootstrap intervals. Filled marks have intervals that exclude zero.}
  \label{fig:contrast}
\end{figure}

\Cref{tab:main} gives the mean scores of every system, \cref{fig:results} shows how single outputs spread around the reference marks, and \cref{fig:contrast} shows what adding each input changes.
Together they support three findings. Every system draws the scenes well, but the systems fail on different requirements once a scene must be spread over separate pieces. A layout helps the pieces fit, while a reference helps neither the count nor the fit. What every system fails at is giving each piece its own part of the scene, and it fails even without tabs and sockets.

\paragraph{Each system fails in its own way.}
The whole-scene controls tell a system that cannot draw a scene from one that cannot split it (\cref{sec:task}). For all five systems these controls score close to the exact cuts of the photographs on every question, so every shortfall below arises only when the scene must be spread over pieces.
All five systems draw separate pieces, though they miscount in different ways. GPT Image 2.6 almost always draws exactly $N$. Nano Banana 2.1 and Muse Image 1.0 draw extra pieces, increasingly so as $N$ grows. Muse Image 1.0 also lets pieces run off the canvas, mostly at four pieces. Qwen-Image-2.1 more often draws too few, joining neighbors into larger regions. FLUX 3 Image Klein errs in both directions, and at sixteen pieces about half of its outputs join some neighbors.
Interlock fails far more often. GPT Image 2.6 interlocks best, yet at nine and sixteen pieces fewer than half of its outputs reach the score of an exact cut whose tabs have slid a tenth of a cell along their edges. Nano Banana 2.1 almost never reaches it at sixteen. FLUX 3 Image Klein interlocks worst, although at four and nine pieces most of its outputs have the right count. Seams left straight on both sides add nothing to this score, yet the two systems that draw them most often also interlock worst (appendix).
Reconstruction fails for every system. No output of GPT Image 2.6, Nano Banana 2.1, Muse Image 1.0 or FLUX 3 Image Klein restores its reference as well as an exact cut whose content is turned only $5^\circ$ inside every piece. The typical output of the two strongest systems sits near pieces whose content is shrunk to half scale.
Roughly two outputs in five of GPT Image 2.6 and Nano Banana 2.1 score no higher than an exact cut of another scene, and so do most outputs of Qwen-Image-2.1 and FLUX 3 Image Klein.
The pieces thus do not carry the reference cell by cell. Nor are they miniatures of the scene, since the diagnostic that compares each piece with the whole reference stays far below the level of thumbnail pieces. Reconstruction drops by only a few hundredths more when the pieces must also fit by shape in the same arrangement, so scoring the two apart hides no conflict between outline and content (appendix).
Precision and recall separate two further errors: the extra pieces of Nano Banana 2.1 at sixteen lower precision, and the merged pieces of Qwen-Image-2.1 lower recall.

\paragraph{A layout helps the pieces fit, and a reference does not.}
Adding each input in turn shows which help the systems can use. With a layout, four of the five systems interlock better (\cref{fig:contrast}), most of all Qwen-Image-2.1, whose pieces rarely fit without one and fit far better with one. This suggests that these systems can copy matching outlines when shown them but seldom create them. Reconstruction rises only slightly, if at all.
A reference shows the very picture to be cut, yet it makes no system scatter or interlock measurably better, and three get worse. Qwen-Image-2.1 returns too few pieces markedly more often, and the pieces of Qwen-Image-2.1, FLUX 3 Image Klein and, slightly, Muse Image 1.0 fit worse.
Both inputs thus act on shape, and neither brings the content of the reference into the right pieces.

FLUX 3 Image Klein is the one system that a layout does not help, most likely because of its prompt upsampler. Its generator sees only a caption that its prompt upsampler writes from the prompt and the images, and the captions often describe a tidier puzzle than the prompt. In at least two of every five jobs in each condition they arrange the pieces in rows or a grid, which no prompt requests. With a layout, more than half drop the quarter turns that every prompt asks for, and about half still describe a tidy grid (appendix).
The model card notes that the caption, not the input images, fixes the composition~\cite{bfl2026flux3image}. A layout does help FLUX.2 [klein], a smaller model of the same developer without this stage, on every score (appendix).

\paragraph{Placing content is the hardest part.}
Weak reconstruction could come from the outlines, because content must also fill the tabs, or from assigning parts of the reference to pieces.
To separate the two, we repeated the nine-piece condition with a reference using square tiles in place of jigsaw pieces, once scattered and turned as in the main task and once upright in their original places (appendix).
Without tabs and sockets, every system brings its pieces closer to the reference. Yet even upright tiles in their original places stay much nearer to an exact cut of another scene than to an exact cut of the reference.
The systems can, however, reproduce the reference. Redrawn whole and cut into the same tiles, it comes back markedly closer for every system (appendix).
The hard part is therefore giving each piece its own part of the reference, not reproducing the reference or drawing its outline.

\section{Discussion}
\label{sec:discussion}

Read separately, the three scores give each system its own profile.
GPT Image 2.6 is the most reliable at scatter and interlock, yet its pieces restore the reference no better than those of Nano Banana 2.1.
Nano Banana 2.1 and Muse Image 1.0 lose count as the pieces multiply, a weak point that compositional benchmarks also report~\cite{ghosh2023geneval}. Qwen-Image-2.1 interlocks its pieces only when a layout shows where they go.
Across all five systems, the generators copy the outlines of a layout but not the content of a reference, even into square tiles that keep their places. What fails is giving each piece its own region of the picture.

\paragraph{Scope and outlook.}
\bench measures what a generator puts on the canvas, not how it represents the scene internally. Probing how a model links the regions of one scene that it places far apart is a natural next step.
The input contrasts show where an input may be lost, but for a closed system they cannot tell a rewriting stage from a generator that ignores the image.
Without a reference, the picture exists only in the pieces. Asking whether they still form one continuous scene then calls for a reference-free measure, for instance the share of neighbors that are mutual best matches, which square-jigsaw solvers use without ground truth~\cite{pomeranz2011fully}.
Comparing content families needs more scenes per family.
Dedicated jigsaw samplers~\cite{geng2024visual,chew2026whimsy} can be scored the same way.

\section{Conclusion}
\label{sec:conclusion}
We present \bench, which uses jigsaw generation and separately scored requirements to test whether image generators can keep a picture continuous where its parts are not adjacent on the canvas. Current generators draw coherent scenes when a scene is whole, but once it is broken into separated pieces, the pieces seldom interlock unless their outlines are shown, and none carries its own part of a reference, even when the pieces keep their arrangement. Keeping a picture continuous across space that the canvas no longer shows remains an open problem for image generation.

\section*{LLM Usage Statement}
\label{sec:llm-statement}
\pdfbookmark[1]{LLM Usage Statement}{bm:llm-statement}

\noindent The authors used Claude Opus 5.5 and GPT-6 Astra to assist with coding and writing.
The assistance covered the benchmark software (evaluation, generation runners and analysis scripts), analysis of candidate evaluation measures, the search and screening of openly licensed photographs and the drafting of scene briefs, literature search and citation checking, and drafting and editing the manuscript, its tables and its figures.
The evaluation method includes no language or vision--language model.
The authors directed all work and take full responsibility for the content, code and citations.

{
    \small
    \bibliographystyle{ieeenat_fullname}
    \bibliography{refs}
}
\clearpage
\appendix
\providecommand{\gjbsupp}[1]{\csname @@input\endcsname generated/supp/#1.tex }
\section{Evaluation Details}
\label{sec:supp-evaluation}

This section gives every step and constant of the evaluation in \cref{sec:evaluation}, the full set of constructions behind \cref{tab:anchors}, and the status of every interlock solve.
\Cref{tab:supp-constants} lists the constants with their sources.

\subsection{Readings}
\label{sec:supp-readings}

The background color $b$ is the per-channel median of the outermost one-pixel ring of the image. Every pixel receives the largest absolute channel difference from $b$, clipped to $[0,255]$ and rounded to eight bits, and the triangle method~\cite{zack1977automatic} picks a threshold $t$ on the histogram of these differences.
Pixels at or below $t$ that are 4-connected to the ring form the background. The pieces are the 8-connected components~\cite{rosenfeld1970connectivity} of the remaining pixels, and a component whose largest Euclidean distance to the background is at most two pixels is discarded as a speck.
An image whose ring holds no pixel at or below $t$ has no background and no pieces.
Reading A is this procedure. Reading B applies a $3\times3$ median filter first. Reading C handles shadows: a pixel whose summed intensity is below that of $b$ is first scaled to the summed intensity of $b$ and then compared with $b$, so that a darker pixel of the background's hue reads as background. Reading D applies both B and C.
RQ1 keeps the reading with the count closest to $N$, breaking ties by fewer merged pieces and then by fewer pieces cut by the border, and RQ2 and RQ3 keep the reading with the highest score. Remaining ties go to the earlier reading in the order A, B, C, D.

The same rule applies to every system. On generated outputs the best reading raises scores by more than on constructions and controls, and by different amounts for different systems, so \cref{tab:supp-readings} gives the scores under each reading used alone.
The order of the systems on interlock is the same under all four readings at every piece count.
On reconstruction, GPT Image 2.6 and Nano Banana 2.1 take the first two places under every reading.
The exact count depends most on the reading: readings C and D, which take darker pixels of the background's hue as background, lower it for each system on average over piece counts, and GPT Image 2.6 keeps the highest count under each.

\begin{table*}[t]
\centering
\footnotesize
\setlength{\tabcolsep}{4pt}
\begin{tabular}{@{}llccccccccc@{}}
\toprule
 & & \multicolumn{3}{c}{Exact count (RQ1)} & \multicolumn{3}{c}{Interlock (RQ2)} & \multicolumn{3}{c}{Reconstruction (RQ3)} \\
\cmidrule(lr){3-5}\cmidrule(lr){6-8}\cmidrule(l){9-11}
System & Reading & 4 & 9 & 16 & 4 & 9 & 16 & 4 & 9 & 16 \\
\midrule
\gjbsupp{tab_v3_readings_body}
\bottomrule
\end{tabular}
\caption{\textbf{Scores under each reading used alone.} Means over all four conditions, and over the two conditions with a reference for reconstruction, by piece count. A: threshold only; B: median filter first; C: chromaticity for pixels darker than the background; D: B and C. The interlock score of a reading includes its 600-second solve.}
\label{tab:supp-readings}
\end{table*}

\subsection{Scatter}

With $k$ pieces of areas $a_1,\dots,a_k$, the expected piece area is $\bar a=\sum_i a_i/N$, and piece $i$ counts as $\max(0,\operatorname{round}(a_i/\bar a)-1)$ merged pieces.
A piece is cut by the border when it touches the outermost ring.

\subsection{Piece geometry}

Each piece is cropped with a margin, its mask is turned upright by the projection-profile skew estimate~\cite{bloomberg1995measuring}, and the image is resampled with the same rotation.
The core is the axis-aligned square bounded by the four straight lines through the piece's sides, found on the upright mask, and its side length sets the scale of the piece.
For a side currently on top, the tab is the piece material above the core's top line within the core's column span, and the socket is the background inside the core's top quarter, the triangle between the top corners and the core's center, that connects to the top line.
Both are point sets $(u,h)$, with $u$ the column offset from the side's midpoint and $h$ the depth above or below the line. The other three sides are read the same way after quarter turns.

\subsection{Interlock}

Two sides placed against each other meet at their midpoints. Their fill $F_{ab}$ counts the points where the tab of side $a$ coincides with the mirrored socket of side $b$ plus the points where the tab of $b$ coincides with the mirrored socket of $a$, and two sides of one piece never fill each other.
Binary variables $x_{cpr}$ place piece $p$ in cell $c$ turned by $r$ quarter turns. Every cell holds at most one piece and every piece at most one cell, with equality on the side with fewer elements, so that $\min(k,N)$ pieces are placed.
For each interior seam $\sigma$ between cells $c$ and $c'$ and each pair of sides with $F_{ab}>0$, a continuous variable $z_{\sigma ab}\in[0,1]$ may be positive only when side $a$ faces $c'$ from $c$ and side $b$ faces $c$ from $c'$:
\begin{equation}
  \sum_{b} z_{\sigma ab}\le x_{c,p(a),r_\sigma(a)},\qquad \sum_{a} z_{\sigma ab}\le x_{c',p(b),r'_\sigma(b)},
\end{equation}
where $r_\sigma(a)$ is the turn that brings side $a$ to the face of $c$ toward $c'$, and $r'_\sigma(b)$ the turn that brings side $b$ to the face of $c'$ toward $c$.
The program maximizes $\sum_{\sigma,a,b}F_{ab}z_{\sigma ab}$, and RQ2 is twice this optimum divided by the total tab and socket material of all sides of all pieces.
HiGHS~\cite{huangfu2018parallelizing}, called through SciPy, solves it with its default optimality tolerance.
A solve first receives 60 seconds. Every reading whose solve stopped at that budget, and whose bound could exceed the best score of the output's other readings, was solved again with 600 seconds, and the better arrangement was kept.
\Cref{tab:supp-solver} reports the share of outputs whose best reading was solved to proven optimality and the median relative gap of the others. Outputs whose solve stopped early are scored by their best arrangement found, so their RQ2 can only be underestimated.
At four and nine pieces the best reading of every output is solved to proven optimality. At sixteen pieces, \cref{tab:supp-bound} sets the mean interlock score found beside the mean it would reach if every solve that stopped early attained the bound it reports.
No arrangement can score above that bound. The difference is at most .03 for every system, and the order of the systems is the same under both.

\begin{table}[t]
\centering
\footnotesize
\setlength{\tabcolsep}{3pt}
\begin{tabular}{@{}lcccc@{}}
\toprule
System & opt. & mean gap & RQ2 found & RQ2 bound \\
\midrule
\gjbsupp{tab_v3_bound_body}
\bottomrule
\end{tabular}
\caption{\textbf{Interlock at sixteen pieces against its upper bound.} Opt.: share of outputs whose best reading was solved to proven optimality; mean gap: mean relative gap of the rest. RQ2 bound takes, for every output, the highest bound over its readings.}
\label{tab:supp-bound}
\end{table}

\begin{table}[t]
\centering
\footnotesize
\setlength{\tabcolsep}{3pt}
\begin{tabular}{@{}lcccccc@{}}
\toprule
 & \multicolumn{2}{c}{$N=4$} & \multicolumn{2}{c}{$N=9$} & \multicolumn{2}{c}{$N=16$} \\
\cmidrule(lr){2-3}\cmidrule(lr){4-5}\cmidrule(l){6-7}
System & opt. & gap & opt. & gap & opt. & gap \\
\midrule
\gjbsupp{tab_v3_solver_body}
\bottomrule
\end{tabular}
\caption{\textbf{Interlock solves.} Share of outputs whose best reading was solved to proven optimality, and the median relative optimality gap of the rest. --: every output solved to optimality.}
\label{tab:supp-solver}
\end{table}

\subsection{Reconstruction}

The reference has side $S=1024$ and cell side $c=S/n$.
For a piece with core side $w$ (the mean of its width and height), the reference is blurred with a Gaussian of standard deviation $\sqrt{s^2-1}/2$, where $s=c/w$ is the shrink factor (no blur when $s\le1$), and resampled bilinearly so that the core covers its cell and the tabs cover the neighboring cells, with the reference reflected beyond its border.
Outside the piece's mask the window takes the piece's own pixels.
MS-SSIM~\cite{wang2003multiscale} uses five scales with weights $0.0448,0.2856,0.3001,0.2363,0.1333$, the $11\times11$ Gaussian window with $\sigma=1.5$ and the constants $K_1=0.01$ and $K_2=0.03$ of SSIM~\cite{wang2004image}, halves the image between scales by area averaging, and averages each map over the pixels of the mask at that scale.
Negative contrast-structure and luminance terms are set to zero, as common implementations do, and the score is the mean over the three color channels.
Each piece is compared with each cell in its four quarter turns, and the best turn gives $s_{ij}$. A piece whose core is shorter than eight pixels on a side has too few pixels to compare and keeps $s_{ij}=0$.
The Hungarian method~\cite{kuhn1955hungarian} pairs pieces and cells to maximize $\sum s_{ij}$, and \cref{eq:rq3} turns the paired similarities into the score.
The thumbnail diagnostic compares each piece, in each quarter turn, with the whole reference shrunk onto its core in the same way and keeps the best turn.

\begin{table*}[t]
\centering
\footnotesize
\begin{tabular}{@{}lll@{}}
\toprule
Constant & Value & Source \\
\midrule
Background color & median of the outermost ring & task: plain background \\
Foreground threshold & triangle method & \cite{zack1977automatic} \\
Speck depth & 2 pixels & too thin to be a piece \\
Median filter (readings B, D) & $3\times3$ & smallest symmetric window \\
Merged pieces & area $/\,\bar a$, rounded, minus one & plain reading of the count \\
Interlock score & Dice & \cite{dice1945measures} \\
Solver budget & 60\,s, then 600\,s & computation only \\
Cell side & $1024/n$ & reference size \\
MS-SSIM weights, window, $K_1,K_2$ & published values & \cite{wang2003multiscale,wang2004image} \\
Anti-alias width & $\sqrt{s^2-1}/2$ & removes detail finer than the piece's pixels \\
Smallest comparable core & 8 pixels & too few pixels to compare \\
Reconstruction score & PQ form, no threshold & \cite{kirillov2019panoptic} \\
\bottomrule
\end{tabular}
\caption{\textbf{Constants of the evaluation.} None is fit to generated outputs.}
\label{tab:supp-constants}
\end{table*}

\subsection{Alternatives not used}
\label{sec:supp-alternatives}

Several common choices were considered and not used, for the reasons below.
\begin{itemize}\itemsep1pt
\item \textbf{One pass/fail verdict over the requirements in turn.} A failure at the first requirement hides every later property of the output, so no question gates another.
\item \textbf{Learned models as evaluators}, segmentation models and vision--language judges. In a pilot they missed most pieces and failed to tell exact puzzles from altered ones (\cref{sec:supp-pilot}).
\item \textbf{Semantic embeddings} such as CLIP, DINO or DreamSim for reconstruction. They are largely insensitive to the rotation and scale of content and reward recognizing the scene, which RQ3 must separate from restoring it.
\item \textbf{Otsu's threshold} in place of the triangle method. Piece colors spread the foreground mode, and Otsu's split breaks pieces into fragments.
\item \textbf{Morphological opening} as a reading. It can split two touching pieces and so repair the very error that RQ1 counts.
\item \textbf{Pairing tabs and sockets anywhere} in place of one arrangement. A shifted tab then finds a matching shifted socket on some other piece, and such cuts score like exact ones (\cref{sec:evaluation}).
\item \textbf{Merged pieces against the median piece area.} Small fragments shrink the median, so one large region reads as many merged pieces. All piece material divided by $N$ is unaffected by fragments.
\item \textbf{A matching threshold} as in panoptic quality. Weighting every pair by its similarity lets the score fall gradually with the severity of an error.
\end{itemize}

\subsection{Constructions}
\label{sec:supp-constructions}

Each construction renders the benchmark's own layout of a scene with pieces cut from that scene's reference, changing one property in a known way, for the eight scenes geometry-01, geometry-05, lines-01, lines-03, multi-01, multi-04, parts-01 and parts-03.
Content changes act inside every piece's outline: a turn about the piece center, a scale about it, a cut from a zoomed part of the scene, the whole scene shrunk onto the core (thumbnail), or the content of another scene.
Piece changes remove, duplicate or replace whole pieces. Outline changes keep the content but exchange outlines among the pieces of one cut, or give each piece the outline of a different random cut.
The other scene is the scene three places further in the list above, counting cyclically.
\Cref{tab:supp-constructions} gives every construction on these eight scenes at each $N$. The tab shifts of \cref{tab:anchors} are computed separately, on three scenes: they move every tab and socket along its side by a tenth of the cell. Shifts by 5\% and 20\% of the cell change the scores by at most .02 at $N=9$ and $16$ and by up to .10 at $N=4$.

\begin{table*}[t]
\centering
\footnotesize
\setlength{\tabcolsep}{4pt}
\begin{tabular}{@{}lcccccccccccc@{}}
\toprule
 & \multicolumn{3}{c}{RQ2} & \multicolumn{3}{c}{RQ3} & \multicolumn{3}{c}{Precision} & \multicolumn{3}{c}{Recall} \\
\cmidrule(lr){2-4}\cmidrule(lr){5-7}\cmidrule(lr){8-10}\cmidrule(l){11-13}
Construction ($N$) & 4 & 9 & 16 & 4 & 9 & 16 & 4 & 9 & 16 & 4 & 9 & 16 \\
\midrule
\gjbsupp{tab_v3_constructions_body}
\bottomrule
\end{tabular}
\caption{\textbf{Every construction.} Means over eight scenes, best reading for each question. Content changes leave the outlines, and so RQ2, unchanged. Outline changes leave the content, and so RQ3, unchanged.}
\label{tab:supp-constructions}
\end{table*}

\providecommand{\gjbsupp}[1]{\csname @@input\endcsname generated/supp/#1.tex }
\section{Additional Results}
\label{sec:supp-results}

\Cref{tab:supp-main-ci} repeats \cref{tab:main} with 95\% intervals, \cref{tab:supp-contrast-ci} gives the values plotted in \cref{fig:contrast}, and \cref{tab:supp-cells} gives every measure in every condition.
FLUX.2 [klein] 4B~\cite{bfl2026flux2klein} shows a failure that no main system shows to this degree: nearly all of its outputs contain one region several pieces large, into which all but about one of the requested pieces merge, so it draws one picture with jigsaw-like lines instead of scattered pieces.
Intervals come from a bootstrap over scenes, and each contrast pairs outputs of the same system, scene and piece count.
In \cref{tab:supp-cells}, exact, too few and too many are shares of outputs, merged is pieces per output and border is the share of pieces cut by the image border, the two ways a count hides an error. P and R are the precision and recall of reconstruction, and thumbnail is the diagnostic that compares each piece with the whole reference. Conditions without a reference have no reconstruction.
Precision and recall separate the two ways reconstruction fails. Recall below precision means missing material, as for FLUX.2 [klein] 4B, whose merged regions each match only one cell, and precision below recall means extra pieces, as for Nano Banana 2.1 at sixteen pieces.
The thumbnail diagnostic stays between .29 and .54 in every condition, far below the .92 to .93 it reaches on thumbnail constructions, so no system answers by drawing the scene in miniature.

\begin{table}[t]
\centering
\footnotesize
\setlength{\tabcolsep}{2.5pt}
\begin{tabular}{@{}llccc@{}}
\toprule
System & $N$ & Exact count & Interlock & Reconstruction \\
\midrule
\gjbsupp{tab_v3_main_ci_body}
\bottomrule
\end{tabular}
\caption{\textbf{Main results with 95\% intervals}, without the control row. Exact count and interlock pool the four conditions (128 outputs per row), reconstruction the two conditions with a reference (64).}
\label{tab:supp-main-ci}
\end{table}

\begin{table*}[t]
\centering
\footnotesize
\setlength{\tabcolsep}{4pt}
\begin{tabular}{@{}lccccc@{}}
\toprule
 & \multicolumn{3}{c}{Adding a layout} & \multicolumn{2}{c}{Adding a reference} \\
\cmidrule(lr){2-4}\cmidrule(l){5-6}
System & Exact count & Interlock & Reconstruction & Exact count & Interlock \\
\midrule
\gjbsupp{tab_v3_contrast_ci_body}
\bottomrule
\end{tabular}
\caption{\textbf{Effects of a layout and a reference with 95\% intervals.} Mean paired difference over scenes and piece counts. The layout's effect on reconstruction compares the two conditions with a reference. A reference has no reconstruction column, since reconstruction needs a reference in both conditions compared.}
\label{tab:supp-contrast-ci}
\end{table*}

\begin{table*}[p]
\centering
\scriptsize
\setlength{\tabcolsep}{3.2pt}
\renewcommand{\arraystretch}{0.95}
\begin{tabular}{@{}llcccccccccccc@{}}
\toprule
 & & & \multicolumn{5}{c}{Scatter (RQ1)} & Interlock & \multicolumn{4}{c}{Reconstruction (RQ3)} \\
\cmidrule(lr){4-8}\cmidrule(lr){9-9}\cmidrule(l){10-13}
System & Condition & $N$ & exact & too few & too many & merged & border & RQ2 & RQ3 & P & R & thumbnail \\
\midrule
\gjbsupp{tab_v3_cells_body}
\bottomrule
\end{tabular}
\caption{\textbf{Every measure in every condition}, means over 32 scenes (columns explained in the text).}
\label{tab:supp-cells}
\end{table*}

\subsection{Captions of FLUX 3 Image Klein}
\label{sec:supp-captions}
FLUX 3 Image Klein's generator receives only the caption that its prompt upsampler writes from the prompt and the condition images, and every record keeps this caption.
\Cref{tab:supp-captions} counts three plain word matches in the captions.
The first match counts a caption as arranging the pieces in rows or a grid when a placing word (``arranged'', ``placed'', ``laid out'', ``presentation'' and similar) is followed within the sentence by a grid word (``grid'', ``row'', ``3-by-3'' and similar). No prompt contains such a phrase: without a layout the prompt asks for random placement, and with one it asks for the scattered places of the layout image.
The second counts it as keeping the turns when it names a quarter turn of 90, 180 or 270 degrees, as every prompt does.
The third counts it as keeping random placement when it contains the word ``random''.
The counts are word matches, not a reading of the captions, and serve to show what the generator was told.
In one caption with a layout, for instance, the upsampler takes the layout image as ``the exact placement guide for nine equal square pieces arranged in a regular 3-by-3 grid'' and keeps ``every piece aligned to 0 degrees''.

\begin{table}[t]
\centering
\footnotesize
\setlength{\tabcolsep}{3pt}
\begin{tabular}{@{}lcccc@{}}
\toprule
Condition & Captions & Rows or grid & Turns kept & ``Random'' kept \\
\midrule
\gjbsupp{tab_flux3_captions_body}
\bottomrule
\end{tabular}
\caption{\textbf{What the prompt upsampler of FLUX 3 Image Klein tells its generator.} Captions: number of captions. The other columns give the share of captions with each word match, over all piece counts. --: prompts with a layout do not ask for random placement. The prompt's own ``cut along a regular 3x3 grid'' names no placement and is not a match. Five text-only jobs returned no image and have no caption.}
\label{tab:supp-captions}
\end{table}

\subsection{Square tiles}
\label{sec:supp-squares}
Reconstructing a reference with jigsaw pieces asks for two things at once, outlines that fit and the right part of the picture in each piece. To score the second on its own, two further conditions replace the jigsaw pieces by square tiles. Both give the reference as Image~1, use nine pieces and no layout, and keep the scenes, briefs, output size and settings of the main study, with one output per job for each of the five systems.
Their prompts keep the main prompt's sentences with ``square tiles'' for ``jigsaw pieces'' and ``so every side of every tile is straight'' for the description of tabs and sockets. In the scattered condition the tiles are, as in the main task, ``rotated 0, 90, 180 or 270 degrees'' and placed ``at random across the gray background''. In the in-place condition each tile ``stays upright'', and the prompt ends ``Keep the tiles in their original 3x3 arrangement, with a narrow gap of gray background between neighboring tiles.'' This condition asks for the simplest assignment of content, a split of the reference that moves and turns nothing.
Scatter and reconstruction are scored by the same code as in the main study. Square tiles have no tabs, so interlock is not scored.

The two shapes put RQ3 on different scales. Because the tabs of a jigsaw piece are compared with the neighboring cells, the content of another scene scores higher on jigsaw pieces than on square tiles. Exact cuts of another scene score .43 as jigsaw pieces and .27 as square tiles, and exact cuts of the reference score .92 and .97.
Comparisons across the two shapes therefore use each output's position between the two marks of its shape, $(\text{RQ3}-\text{another})/(\text{exact}-\text{another})$, which is 0 at an exact cut of another scene and 1 at an exact cut of the reference.

\Cref{tab:supp-squares} gives the results.
Without tabs and sockets the position rises for every system, and the interval of the change excludes zero for each.
Keeping the tiles upright in place raises reconstruction further for Nano Banana 2.1, and slightly for Muse Image 1.0.
Even so, no system brings its tiles more than about a third of the way from another scene to an exact cut in either condition, and the tiles of Qwen-Image-2.1 stay at the level of another scene.
Giving each piece its own part of the reference thus remains the main difficulty when the outlines are as simple as possible and the tiles do not move.

A last condition asks whether the systems can reproduce the reference at all. Each system receives the reference as Image~1 with the whole-scene prompt of the main study followed by ``Image 1 shows the scene exactly.'', and the program cuts the returned picture, resized to $1024\times1024$, into the same upright tiles in their original places as the exact cut. This condition differs from the in-place tiles only in who divides the picture.
Drawn whole, the reference comes back far closer: GPT Image 2.6, Muse Image 1.0 and FLUX 3 Image Klein reach about four fifths of the way from another scene to an exact cut, and Nano Banana 2.1 and Qwen-Image-2.1 about half, against at most a third for the tiles that any system draws itself.
Every system loses reconstruction when it must divide the picture itself, and the interval of the loss excludes zero for each.
The difficulty thus lies in dividing the picture, not in reproducing it.

The in-place prompt asks for a narrow gap, and GPT Image 2.6 often leaves gaps a few pixels wide or filled with shadow, which the scatter score reads as tiles joined into one piece. Under the main study's rules this counts against the output, and it lowers both the count and reconstruction. Among its in-place outputs with exactly nine separate tiles, the mean position is .27, still far from an exact cut.
One FLUX 3 Image Klein job in the in-place condition returned no image, for the reason given in \cref{sec:supp-protocol}, and counts as an output with no pieces.

\begin{table*}[t]
\centering
\footnotesize
\setlength{\tabcolsep}{3.5pt}
\begin{tabular}{@{}lcccccccccc@{}}
\toprule
 & \multicolumn{3}{c}{Exact count (RQ1)} & \multicolumn{4}{c}{Reconstruction (RQ3)} & \multicolumn{3}{c}{Change} \\
\cmidrule(lr){2-4}\cmidrule(lr){5-8}\cmidrule(l){9-11}
System & Jigsaw & Scattered & In place & Jigsaw & Scattered & In place & Whole & \shortstack{Position,\\scattered $-$ jigsaw} & \shortstack{RQ3, in place\\$-$ scattered} & \shortstack{RQ3, whole\\$-$ in place} \\
\midrule
\gjbsupp{tab_squares_body}
\bottomrule
\end{tabular}
\caption{\textbf{Square tiles in place of jigsaw pieces}, nine pieces, with the reference and no layout. Jigsaw: the main study's outputs in that condition. Scattered: square tiles turned and placed at random. In place: upright square tiles in their original arrangement. Whole: the reference reproduced as one picture and cut by the program into in-place tiles, so its count is exact by construction and not shown. Means over 32 scenes. The changes are mean paired differences over scenes, bold where the 95\% scene-cluster bootstrap interval excludes zero. The position change compares the scattered tiles with the jigsaw pieces on the scale of each shape's marks (text). The last two rows give the marks of each shape.}
\label{tab:supp-squares}
\end{table*}

\begin{table*}[!t]
\centering
\footnotesize
\setlength{\tabcolsep}{3.5pt}
\begin{tabular}{@{}lcccccccccccc@{}}
\toprule
 & \multicolumn{3}{c}{RQ3, best pairing} & \multicolumn{3}{c}{RQ3, one assembly} & \multicolumn{3}{c}{Connectors present} & \multicolumn{3}{c}{Connectors fitted} \\
\cmidrule(lr){2-4}\cmidrule(lr){5-7}\cmidrule(lr){8-10}\cmidrule(l){11-13}
 & 4 & 9 & 16 & 4 & 9 & 16 & 4 & 9 & 16 & 4 & 9 & 16 \\
\midrule
\gjbsupp{tab_joint_body}
\bottomrule
\end{tabular}
\caption{\textbf{Shape and content in one assembly}, by piece count. RQ3 with the best pairing of pieces and cells, as in the main study but on the reading used for interlock, and in the one arrangement that fits the most seams and then matches the reference best. Systems: means over the two conditions with a reference for RQ3 and over all four conditions for the connector shares, with a job that returned no image counted as no pieces. Whole-scene control: the program's exact cut of each system's whole picture. Constructions: means over the eight construction scenes. Square tiles: the exact in-place cut of the reference into nine upright tiles.}
\label{tab:supp-joint}
\end{table*}

\subsection{Shape and content in one assembly}
\label{sec:supp-joint}
Interlock and reconstruction each take the arrangement of the pieces that suits them best, the interlock arrangement and the best pairing of pieces with cells, while a real jigsaw has to satisfy both in one assembly. The interlock score also leaves out a seam where neither side carries a tab or socket, since such a seam adds no material, so it does not show how many seams carry connectors at all. To address both points, every output is assembled once more in a single arrangement.

This arrangement first fits as many interior seams as possible and then, among the arrangements that do, puts in each cell the piece whose content best matches that cell of the reference. A seam counts as \emph{fitted} when the tab on one side and the socket on the other overlap with intersection over union above one half, the matching rule of panoptic quality~\cite{kirillov2019panoptic}. Both goals are solved as one integer program with the same variables and solver as interlock, in which each fitted seam outweighs all similarities together. The arrangement gives three scores. \emph{RQ3 in one assembly} compares each placed piece, in its placed turn, with its cell, as RQ3 does, and counts a piece left out of the grid as unpaired, so it asks, like direct comparison~\cite{cho2010probabilistic}, whether each piece shows its own cell where its outline puts it. \emph{Connectors present} is the share of the $2n(n-1)$ interior seams at which either side carries a tab or socket, and \emph{connectors fitted} is the share of fitted seams. Each output is read as for its interlock score. Every program at four and nine pieces was solved to optimality, and so were nine in ten at sixteen pieces. The rest stopped at the 600\,s limit with their best arrangement found.

The constructions of \cref{tab:supp-joint} show what the arrangement adds. An exact cut and the whole-scene control keep their reconstruction in one assembly. When the outlines are exchanged within the cut, every seam still fits and the best pairing still finds the right content, but in one assembly reconstruction falls most of the way to another scene, because the outlines now place each piece in the wrong cell. Scored apart, interlock and reconstruction cannot see this conflict. With outlines from unrelated cuts, the loss is smaller, and at four pieces there is none, because only one seam in four fits and content decides the arrangement. All cuts share one tab outline and differ only in which side carries the tab, so a fitted seam shows that a tab meets a socket of its size, not that the two were cut together. How closely they fit is left to the interlock score. Square tiles show almost no connector material, as they should.

For the generated outputs, requiring one assembly costs reconstruction only a few hundredths, about as much as it costs an exact cut of another scene, and reconstruction in one assembly stays near that mark and far below the whole-scene control. Content placed against its outlines is therefore not what holds reconstruction down, since even the best pairing finds little of the reference. GPT Image 2.6, Nano Banana 2.1 and Muse Image 1.0 draw tabs or sockets at nearly every seam. Qwen-Image-2.1 and FLUX 3 Image Klein leave both sides straight at one seam in six to ten, yet these two also interlock worst, so leaving out seams without connectors does not lift their interlock scores. Tabs meet sockets of their size at about nine seams in ten for GPT Image 2.6 and Muse Image 1.0, fewer for Nano Banana 2.1 as the pieces grow in number, and at little more than half or less for Qwen-Image-2.1 and FLUX 3 Image Klein. Together with the interlock scores, this shows that the strongest systems usually put a tab against a socket, but seldom make the two fit as closely as a cut does.

\section{Generation Provenance}
\label{sec:supp-protocol}

\label{sec:supp-generation}
\Cref{tab:supp-generation} lists each system's identifier, setting, output size and cost.
Each system produced 256 scattered images at nine and sixteen pieces, 128 at four pieces and 32 whole-scene pictures, except for five text-only jobs of FLUX 3 Image Klein. In each of these, the prompt upsampler reached its limit of 3{,}584 new tokens without finishing the caption. A caption cut off at the limit is incomplete, so these jobs return no image and count as outputs with no pieces. The longest finished caption has 2{,}753 tokens.
The API models have no dated snapshots. Every request logs the requested identifier and its timestamp. Nano Banana 2.1 echoes its identifier in every response, and GPT Image 2.6 and Muse Image 1.0 responses carry none.
No request enabled a tool, and Muse Image 1.0's web search, image search and shell tools were disabled explicitly.
FLUX 3 Image Klein runs its model card's settings: its distilled checkpoint with 16 steps and no guidance, after the required prompt upsampler (\texttt{black-forest-labs/\allowbreak flux-3-image-klein-pup}, revision \texttt{a5789954}), which decodes greedily with thinking off.
Qwen-Image-2.1 runs its pipeline's defaults, which its model card's examples use: 40 steps and true CFG scale 1.0 without a negative prompt, so true classifier-free guidance stays off, as in those examples.
API jobs are sent without a seed, whereas open-weight jobs are seeded from their job identifiers. Each job keeps its first complete response, and every API job completed at its first attempt.
Costs follow the providers' published prices, retrieved on 9 November 2026: token rates for GPT Image 2.6 and Nano Banana 2.1, a flat price per image for Muse Image 1.0.
The open-weight models ran in one session on a single NVIDIA RTX PRO 6000 Blackwell GPU with PyTorch 2.14.1, CUDA 13.0 and Diffusers 0.42.1, at peak allocations of 20\,GB (FLUX.2 [klein] 4B) and 47\,GB (Qwen-Image-2.1), consuming about 16 Google Colab compute units.
FLUX 3 Image Klein ran all its jobs in one session on the same GPU type with PyTorch 2.11.0, CUDA 13.0 and Diffusers 0.43.0, which provides its pipeline, at a peak allocation of 45\,GB including the upsampler, consuming about 19 compute units.
The four-piece jobs of Qwen-Image-2.1 and FLUX.2 [klein] 4B ran with the same software and fixed settings, Qwen-Image-2.1 on the same GPU type and FLUX.2 [klein] 4B on an NVIDIA L4 (24\,GB).
The square-tile jobs of \cref{sec:supp-squares} ran with the same settings: the three API models in one run each, at 11.12, 3.45 and 0.64 USD for GPT Image 2.6, Nano Banana 2.1 and Muse Image 1.0, and each open-weight model in its own session on the same GPU type, together about 7 compute units. The whole reproductions of the reference ran the same way, at 7.02, 1.51 and 0.32 USD for the three API models and about 3 compute units for the two open-weight models.

\begin{table*}[t]
\centering
\footnotesize
\setlength{\tabcolsep}{3.5pt}
\begin{tabular}{@{}p{0.13\linewidth}p{0.22\linewidth}p{0.21\linewidth}lllr@{}}
\toprule
System & Identifier & Setting & Scattered & Whole & Latency & Cost (USD) \\
\midrule
\raggedright FLUX 3 Image Klein~\cite{bfl2026flux3image} & \raggedright\texttt{black-forest-labs/\allowbreak flux-3-image-klein}, revision \texttt{09561231} & \raggedright bfloat16, 16 steps, no guidance, prompt upsampler & $1536\times1024$ & $1024\times1024$ & 15.6\,s & GPU \\
\raggedright FLUX.2 [klein] 4B~\cite{bfl2026flux2klein} & \raggedright\texttt{black-forest-labs/\allowbreak FLUX.2-klein-4B}, revision \texttt{e7b7dc27} & \raggedright bfloat16, 4 steps, guidance 1.0 & $1536\times1024$ & $1024\times1024$ & 1.8\,s & GPU \\
\raggedright Qwen-Image-2.1~\cite{qwen2026qwenimage21} & \raggedright\texttt{Qwen/\allowbreak Qwen-Image-2.1}, revision \texttt{d26bb612} & \raggedright bfloat16, 40 steps, true CFG 1.0, key-value cache & $1536\times1024$ & $1024\times1024$ & 17.7\,s & GPU \\
\raggedright GPT Image 2.6~\cite{openai2026gptimage26} & \raggedright\texttt{gpt-image-2.6-sunburst}, Images API & \raggedright quality max, complexity high, moderation low & $1536\times1024$ & $1024\times1024$ & 92.2\,s & 45.03 / 51.78 \\
\raggedright Gemini Nano Banana 2.1~\cite{google2026nanobanana21} & \raggedright\texttt{gemini-nano-banana-2.1}, Gemini API & \raggedright image size 1K, aspect 3:2, thinking level high & $1264\times848$ & $1024\times1024$ & 34.6\,s & 14.41 / 15.83 \\
\raggedright Muse Image 1.0~\cite{meta2026museimage} & \raggedright\texttt{muse-image-1.0}, Meta model API & \raggedright reasoning strength high, tools off & $1920\times1280$ & $1600\times1600$ & 23.3\,s & 2.56 / 2.88 \\
\bottomrule
\end{tabular}
\caption{\textbf{Generation provenance.} Sizes are as returned for all scattered and whole-scene jobs of each system (scattered jobs request $1536\times1024$, whole-scene jobs a square). Latency is the median over scattered jobs, and for FLUX 3 Image Klein includes the upsampler. Cost covers the nine- and sixteen-piece runs, as the 256 scattered jobs and the 288 images with the whole-scene pictures, not the four-piece jobs. GPU: open-weight models, run on rented GPUs (text). Per image, GPT Image 2.6 cost 0.17--0.19, Nano Banana 2.1 0.04--0.08 and Muse Image 1.0 0.01.}
\label{tab:supp-generation}
\end{table*}

\section{Scenes, Briefs, Prompts and Layouts}
\label{sec:supp-scenes}

\paragraph{Admission rules.}
Every scene is a photograph from Wikimedia Commons.
A candidate was admitted when (i) its license is CC0 with no further restriction, re-queried through the Commons API on 9 November 2026; (ii) a square crop of at least 1024 pixels (the smallest used is 1080) contains the scene, resized to $1024\times1024$ with area interpolation; (iii) the crop shows no text, sign, logo or identifiable person; (iv) it is no near-duplicate of another scene, by 64-bit difference hashes, HSV-histogram distances, $16\times16$ gray-layout correlations over all pairs and a visual check; and (v) its rim stays distinct from the background color $(235,232,224)$ of the layouts and the cut controls: in the outer 32-pixel band, at most 25\% of the pixels overall and 50\% on every edge lie within CIE76 $\Delta E<20$ of it.
Eight scene-geometry sources are Wikimedia standard renditions of 1280 or 1920 pixels.

\paragraph{Curation.}
For each family, an LLM agent searched Commons for CC0 bitmaps, screened candidates against the rules, and proposed eight scenes with alternates, crop boxes and briefs.
A separate agent then cross-checked all 32 scenes: it re-queried every license, reproduced every crop pixel for pixel, viewed every crop at full size with zooms on suspect regions, re-ran the brief checks, compared every number word and relation with the crop, and screened diversity.
It revised 12 briefs whose wording the crop did not support, re-cropped two scenes whose rims came too close to the background color, and replaced two scenes, one repeating the subject of another with a doubtful count and a visible sign, the other with a count that could not be verified.
No curation step opened any model output.

\paragraph{Brief rules.}
A brief has 15--35 words in one to three sentences, ASCII only and no digits; every countable object it names carries a number word giving its total in the scene; relations refer to the upright reassembled picture; no word refers to the puzzle itself (checked against a fixed list such as \emph{piece}, \emph{tab} or \emph{grid}); and the words gray, grey, white, beige and cream never name a region touching the rim.

\paragraph{Prompts.}
Every scattered prompt begins with the same core, shown here for $N=9$ (\texttt{3x3} becomes \texttt{4x4} at $N=16$) with the brief in braces:
\begin{quote}\small\ttfamily
Show only a plain, uniform warm light gray background and exactly 9 jigsaw pieces cut from one scene. When fitted together, they form one continuous scene: \{brief\} The pieces were cut along a regular 3x3 grid: each shared side has one tab and its matching socket, and the outer sides are straight. Each piece shows only its part of the scene, with no outline stroke. Objects may be split across pieces. Pieces lie flat in the image plane at one common scale, each fully surrounded by the background. Each piece, with its part of the scene, is rotated 0, 90, 180 or 270 degrees.
\end{quote}
\noindent The conditions append the text of \cref{tab:cells}.
The whole-scene prompt is
\begin{quote}\small\ttfamily
Show one continuous scene: \{brief\} The scene fills the entire image.
\end{quote}
\noindent Every rendered prompt is checked to be ASCII, one line with single spaces, ending in a period, free of leftover braces, containing the brief exactly once, and naming the right $N$, grid and image numbers. The longest prompt has 162 words.

\paragraph{Layouts.}
The pieces come from one fixed jigsaw cut of a $720\times720$ square, with tabs 0.15 of a cell side deep, drawn at scale 0.6792, so that the whole puzzle spans 489 pixels.
For each scene and $N$, a seed derived from the scene identifier and $N$ draws a permutation of the pieces over the slots, a uniform quarter-turn rotation for each piece, and then integer centers one piece at a time, uniformly over positions that keep the outline at least 12.07 pixels from the canvas edge and at least 35.41 pixels from every placed piece, the smallest outline gap of the benchmark's sixteen-piece example layout.
Every layout was placed at its first draw.
The whole-scene cut controls and reference cuts place their pieces in the scene's layout, so the cut controls and the layout conditions share one geometry.

\section{Pilot: Learned Models as Evaluators}
\label{sec:supp-pilot}

An exploratory pilot asked whether learned models can tell whether pieces form one picture.
It used square pieces with known answers and is separate from the study.

\paragraph{Segmentation.}
SAM~3~\cite{carion2025sam3} (\texttt{facebook/sam3}, revision \texttt{3c879f39}), prompted with the text \texttt{square picture}, kept masks with score above 0.5 and matched them one to one at IoU at least 0.5.
On eight exact puzzles of 9 and 16 square pieces cut from two photographs and scattered in two layouts, it found 29 of the 100 pieces, with no false positive, and returned no mask at all on six of the eight images.

\paragraph{Paired verification.}
Two open vision--language models, \texttt{google/gemma-4-E4B-it} and \texttt{google/gemma-4-31B-it} at fixed revisions, saw the intact source beside each puzzle and answered with a strict JSON verdict under greedy decoding.
Each configuration held four pairs, an exact puzzle and its twin in which one piece is replaced by a copy of another at the same slot and rotation, and a pair counts as told apart only when both verdicts are right.
Eight configurations were run, each a combination of one of the two models, computer-drawn or photographic sources, a conservative or a neutral prompt wording, and an image budget of 260 or 1080 tokens. No configuration told apart more than one of its four pairs, and six of the eight rejected every puzzle, exact ones included.

\end{document}